# Transfer Learning and Class Decomposition for Detecting the Cognitive Decline of Alzheimer's Disease

Maha M. Alwuthaynani, Zahraa S. Abdallah and Raul Santos-Rodriguez

**Abstract** Early diagnosis of Alzheimer's disease (AD) is essential in preventing the disease's progression. Therefore, detecting AD from neuroimaging data such as structural magnetic resonance imaging (sMRI) has been a topic of intense investigation in recent years. Deep learning has gained considerable attention in Alzheimer's detection. However, training a convolutional neural network from scratch is challenging since it demands more computational time and a significant amount of annotated data. By transferring knowledge learned from other image recognition tasks to medical image classification, transfer learning can provide a promising and effective solution. Irregularities in the dataset distribution present another difficulty. Class decomposition can tackle this issue by simplifying learning a dataset's class boundaries. Motivated by these approaches, this paper proposes a transfer learning method using class decomposition to detect Alzheimer's disease from sMRI images. We use two ImageNet-trained architectures: VGG19 and ResNet50, and an entropy-based technique to determine the most informative images. The proposed model achieved state-of-the-art performance in the Alzheimer's disease (AD) vs mild cognitive impairment (MCI) vs cognitively normal (CN) classification task with a 3% increase in accuracy from what is reported in the literature.

Maha M. Alwuthaynani
University of Bristol, UK & College of Computer Science and Information Systems, Najran University, Saudi Arabia e-mail: maha.alwuthaynani@bristol.ac.uk

Zahraa S. Abdallah
University of Bristol, UK e-mail: zahraa.abdallah@bristol.ac.uk

Raul Santos-Rodriguez
University of Bristol, UK e-mail: enrsr@bristol.ac.uk

# 1 Introduction

Dementia is a broad term for various mental pathologies that can cause memory troubles and brain changes. Alzheimer's disease (AD) is the cause of approximately 60–80% of dementia cases. People with Alzheimer's experience many symptoms that change over the years, reflecting the degree of damage to neurons in different parts of the brain. This disease starts years before its symptoms are present, and the speed with which symptoms progress from mild to moderate to severe varies from one individual to another [4].

The use of neuroimaging modalities has been demonstrated to significantly aid in the diagnosis of Alzheimer's disease. Structural magnetic resonance imaging (sMRI) is the most widely used neuroimaging modality for AD detection and has shown increased performance in the literature. Moreover, sMRI is capable of capturing grey matter atrophy related to the loss of neurons and synapses in AD as well as white matter atrophy linked to the loss of integrity of the white matter tract. Therefore, atrophy measured by sMRI is considered a robust AD biomarker [6].

According to [16], machine-learning approaches are valuable for diagnosing Alzheimer's. In addition, the use of deep learning models has become widespread for dealing with medical images. Deep learning has gained considerable interest in Alzheimer's detection research since 2013, and the number of publications on this topic has risen drastically since 2017 [6]. However, there are some limitations when training the model from scratch. The main limitation is that training models demand a significant amount of labelled data. Another limitation of using deep learning on sMRI data is that model training requires a large number of computational resources. It is also challenging to deal with irregularities in the dataset distribution.

Transfer learning is an alternative for training the model from scratch [3]. Transfer learning is an important mechanism in machine learning for addressing the issue of insufficient training data. It attempts to transfer knowledge from the source domain to the target domain [15].

Class decomposition assists with the issue of irregularities in the dataset distribution by making learning the class boundaries of a dataset more uncomplicated. The class decomposition aims to divide each class in the image dataset into subclasses, with each subclass being treated independently, simplifying the dataset's local structure to deal with any irregularities in the data distribution [18].

In this paper, we examine how transfer learning and class decompaction can be applied for enhanced diagnosis of AD. The essential motivation behind utilising transfer learning is tackling the challenges of the lack of availability of a large annotated training set. The contributions of our method can be summarised as follows:

- We proposed an efficient transfer learning-based approach for diagnosing Alzheimer's from sMRI scans.
- Investigated the influence of transfer learning across two different domains ImageNet data and sMRI images.

- We employed the image entropy strategy to select the most informative information for training the model when even using a small training dataset to achieve better performance
- We utilised class decompaction to uncover the hidden patterns within Alzheimer's images by dividing classes into sub-clusters and to overcome irregularities in distribution.

The rest of this paper is organised as follows: Section 2 describes the related work, the methodology is introduced in Section 3, and Section 4 presents our experiments and findings. Finally, Section 5 concludes the work.

## 2 Related works

In this section, we aim to present state-of-the-art on how neuroimaging is utilised to diagnose and monitor Alzheimer's progression using voxel-based and slice-based methods, provide an overview of transfer learning approach and its application in the detection of Alzheimer's disease and explore how the class decomposition method can be used to assist in enhancing models performance.

Many studies documented in the literature have assessed structural brain variances to highlight the atrophy of AD and prodromal AD spatially distributed over many brain regions. In the following sections, we explore how neuroimaging is utilised to diagnose and monitor AD progression using voxel-based and slice-based diagnostic techniques.

Various studies have proposed models that rely on the voxel-based method. These involve voxel-wise analyses of local brain tissue to determine the pathological modifications in discriminative regions for AD diagnosis. The voxel-based method utilises voxel intensity values from whole neuroimaging modalities or tissue components. Each image demand is standardised to a 3D space [6]. The authors of [6] stated that approximately 70% of investigations utilising this approach involve a full-brain analysis. The advantage of a full brain analysis is that the spatial data are completely integrated, which allows for obtaining 3D data from neuroimaging scans. The disadvantage is that it causes increasing amounts of data dimensionality and computational load [2]. Many studies have employed distinct strategies using the voxel-based approach [17], [19].

The slice-based approach is employed to extract two-dimensional (2D) slices from 3D brain scans. As actual brain tissue is represented in a 3D format (3D brain scans), utilising a slice-based approach might result in data loss because it reduces volumetric data to 2D representations [2], [6]. Many investigations have utilised distinctive approaches to extract 2D image slices from 3D brain scans, whereas others have employed standard projections of the axial, sagittal, and coronal planes. However, none of these studies achieved a full-brain analysis because the 3D brain scans could not be converted into 2D slices. Therefore, a whole-brain analysis is not achievable using the slice-based approach [6]. In [2], the authors stated that using a 2D slice strategy decreases network complexity and the parameters required to train the model. At the same time, it has the drawback of spatial dependency loss between nearby slices. Many studies have employed distinct strategies to extract two-dimensional slices from 3D brain scans [9], [12], [10], [8], [14].

Transfer learning is a key mechanism in machine learning for dealing with the issue of insufficient training data [15]. It effectively extends knowledge previously learned in one task to a new task [13]. Another issue that can be addressed using Transfer learning is that many machine learning approaches perform sufficiently only under a standard assumption: the training and testing dataset have the same feature space and distribution. Thus, most statistical models need to be rebuilt from scratch when the distribution changes using newly collected training data which in some cases could be an expensive re-collect the required training data and reconstruct the models. Transfer learning between task domains would be desirable to address these issues and reduce the need and effort to re-collect the training data [13]. According to [3], transfer learning approaches have shown robust performance because it transfers knowledge across domains. Moreover, many research used two transfer learning techniques: (1) employing a pre-trained network as a feature extractor, which is not demanding to train the network at all and (2) fine-tuning a pre-trained network on the data under study [11].

Many studies utilised pre-trained networks on the ImageNet dataset as feature extractors of medical images to overcome the lack of large-scale annotated data [3]. In [12], the authors proposed a transfer-based learning network that predicts Alzheimer's disease using sMRI scans. Authors of [10] suggested a layer-wise transfer learning-based model investigating the relationship between training size and classification accuracy in the context of transfer learning with intelligent data selection. The authors determined the most useful two-dimensional slices taken from three-dimensional sMRI images using an entropy-based method based on calculating the image entropy using a histogram. The model is very similar to the VGG-19 design. The fully connected layers were adjusted in all four configurations to test the model. Each configuration involved periodically freezing some blocks and altering the amount of the training dataset.

The class decomposition method was proposed in 2003 by [18], which is based on using clustering for pre-processing images. It enables the reduction of the impact of noisy data, finds hidden patterns within each class, and improves classification accuracy. The clustering-based class decomposition approach works by applying clustering to samples in a class to split it into sub-classes and then re-labelling each cluster's instances with a new class label [18]. Many studies have utilised class decomposition. Authors of [1] suggested CNN architecture called DeTraC, (Decompose, Transfer, and Compose) for adapting the class decomposition to medical image classification tasks. The suggested model was validated using three distinct datasets: chest X-rays, digital mammograms, and histological sections of human colorectal cancer. To increase the number of samples, data augmentation techniques like flipping, translation (translating, scaling, and rotation at various angles), colour processing, and minor random noise perturbation were used. The authors used the three primary scenarios of shallow-tuning, fine-tuning, and deep-tuning to evaluate

DeTraC with five different pre-trained CNN networks (AlexNet, VGG16, VGG19, GoogleNet, and ResNet)

## 3 Methodology

The architecture of the proposed model is inspired by the [1] work. Our proposed approach uses sMRI scans from the Alzheimer's Disease Neuroimaging Initiative (ADNI) database. Figure 1 illustrates the proposed network using sMRI.

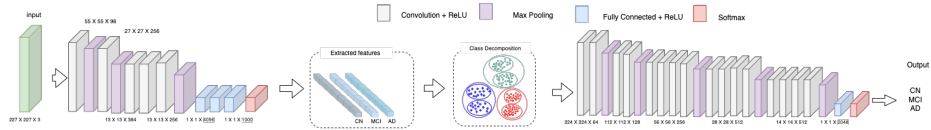

Fig. 1: The architecture of the proposed model consists of AlexNet for feature extraction, class decomposition for splitting classes into sub-classes, and VGG19 network for classification

Images for all subjects go through three phases: feature extraction, clustering, and classification. Transfer learning is used to capture the features from images. In the clustering stage, the data samples of each class are divided into clusters (sub-classes) to feed into the classification phase. All sub-classes classified as $AD_1$, $AD_2$, $MCI_1$, $MCI_2$, $CN_1$ and $CN_2$ are then assembled back to construct the actual classes ($AD$, $MCI$, and $CN$) before the class decomposition process to produce the prediction.

### 3.1 Selection of the Most Informative Training Dataset

Alzheimer's Disease Neuroimaging Initiative (ADNI) database provided the structural magnetic resonance imaging (sMRI) data used in this investigation. Three-dimensional (3D) Neuroimaging Informatics Technology Initiative (NIFTI) formatted sMRI data were used for this study. However, handling 3D images necessitates powerful computing capabilities and a big memory space. Consequently, employing two-dimensional sMRI slices as an alternative to three-dimensional images is one option to lessen processing. Creating 2D slices from 3D sMRI scans produces a vast number of images, some of which contain noisy data while others are rich in information. Selecting the most relevant data is essential to the method's success. Therefore, we employed the image entropy method to select the most informative 2D slices, as opposed to most current techniques, which randomly select the two-dimensional images for training and testing the model. After extracting all two-dimensional slices for each subject, we calculate each slice's image entropy using the grey-level cooccurrence matrix (GLCM) [7]. It is a statistical technique for analysing texture, investigates the spatial relationship between pixels and determines how frequently a combination of pixels appears in an image in a given direction and distance [5]. The two-dimensional images then sort in descending order based on the entropy of the images, and images with the greatest entropy values are the most informative. We pick only twenty slices with the highest entropy from each subject for training and testing the model. The following formula is applied for calculating the image entropy of a set of $M$ symbols with probabilities $p_1, p_2, ..., p_M$:

$$H = -\sum_{i=1}^{M} p_i \log p_i. \tag{1}$$

### 3.2 Feature Extraction

For feature extraction, we use transfer learning that uses a pre-trained model on ImageNet to capture the general features from images by freezing some layers of the pre-trained network and adapting the top layer to be used for AD data. AlexNet network is used to extract features from two-dimensional images. The top layer of the pre-trained network is adopted for the three classes AD, MCI and CN. After extracting the features, we used principal component analysis (PCA) to reduce the dimensionality of the feature space, which assists in reducing memory requirements and enhancing the framework's efficiency. Then, the extracted features were passed to the cluster to perform the class decomposition.

### 3.3 Class Decomposition

Applying clustering as a pre-processing phase for each class is known as class decomposition. This approach was proposed by [18]. The idea of the clustering-based class decomposition approach is that clustering is applied to all data samples of each class to divide the class into clusters (sub-classes) and to re-label each cluster's instances with a new class label. This technique assists in decreasing the impact of noisy data, discovering the hidden patterns within each class and enhancing classification accuracy [18].

Suppose the feature space is illustrated by a two-dimensional matrix ($A$), where $A$ is the image dataset, $L$ is a set of class labels, and $n$, $m$, and $k$ are the number of images, features and classes, respectively. $A$ and $L$ can be written as in (2).

$$A = \begin{bmatrix} a_{1_1} & a_{1_2} & ..... & a_{1_m} \\ a_{2_1} & a_{2_2} & ..... & a_{2_m} \\ . & . & . & . \\ a_{n_1} & a_{n_2} & ..... & a_{n_m} \end{bmatrix}, L = \{l_1, l_2, ...., l_k\} \tag{2}$$

Class decomposition is applied to partition each class in a dataset ($A$) into $k$ sub-classes, where each subclass is treated independently. The new class label will be a pair $(c, k')$, where $c$ denotes the actual (class label) from label space $Y$, and $k'$ represents the (cluster label) to which the sample belongs from the new cluster label space $Y'$. The class decomposition resulted in a new dataset ($B$) with new sub-classes. A and B datasets can be written as in (3).

$$A = \begin{bmatrix} a_{1_1} & a_{1_2} & \dots & a_{1_m} & l_1 \\ a_{2_1} & a_{2_2} & \dots & a_{2_m} & l_1 \\ . & . & . & . & . \\ a_{n_1} & a_{n_2} & \dots & a_{n_m} & l_2 \end{bmatrix}, B = \begin{bmatrix} b_{1_1} & b_{1_2} & \dots & b_{1_m} & l_{1_1} \\ b_{2_1} & b_{2_2} & \dots & b_{2_m} & l_{1_k} \\ . & . & . & . & . \\ b_{n_1} & b_{n_2} & \dots & b_{n_m} & l_{2_k} \end{bmatrix} \tag{3}$$

### 3.4 Classification and Class Composition

In the classification phase, VGG19 was employed for the classification task after the class decomposition, as shown in Figure 1. We have also experimented with other pre-trained models to emphasise the effectiveness and robustness of our proposed method. The top layer of the VGG19 network is adopted for in $Y'$. The classifier was trained on the new dataset ($B$), which was produced after decomposing the classes. The classifier constructs a hypothesis $h'$ and maps samples from class label space $Y$ to the cluster label space $Y'$. The hypothesis $h'(x) = (a, b)$ produces a prediction consisting of a pair, a class label and a cluster label. The cluster label will be removed in the composition phase to obtain the actual prediction in the class label space $Y$. Then, the sub-classes will be reassembled to construct the predicted classes, depending on the dataset before decomposition.

## 4 Results and Discussion

This section provides the experimental results for the model. We executed our deep-learning approach using Keras with a TensorFlow backend. Our target is to differentiate AD subjects from MCI and CN subjects by analysing sMRI scans.

### 4.1 Expieriments setup

sMRI scans used in this study are from the Alzheimer's Disease Neuroimaging Initiative (ADNI) database (available at http://adni.loni.usc.edu). The dataset used in the experiments contains 134 three-dimensional T1-weighted magnetic resonance images, registered using six DOF (rigid-body) to MNI 152 template standard space. Table 1 shows the demographic characteristics of the subjects.

We utilise NiBabel and OpenCV-Python libraries to process NIFTI files and obtain the 2D brain axial plane slices for training and testing the model. In addition, after extracting all two-dimensional slices, we used the scikit-image library to measure the entropy of the images and then picked only twenty slices with the highest entropy from each subject. The dataset is divided as follows: 80% of the subjects were randomly selected for training and validating the model, while the remaining 20% of the subjects were reserved for classifier testing. Figure 2 shows sample slices from the ADNI Dataset across the three classes.

| Characteristic | CN | MCI | AD |
|---|---|---|---|
| Subjects | 44 | 45 | 45 |
| Age range | 61.1- 89.7 | 60.4 – 87.4 | 62.7 – 89.7 |
| Gender (M/F) | 22/ 22 | 26 / 19 | 21/ 24 |
| MMSE range | 27 - 30 | 19 - 30 | 10 - 28 |

Table 1: Demographic characteristics of subjects for the ADNI sample

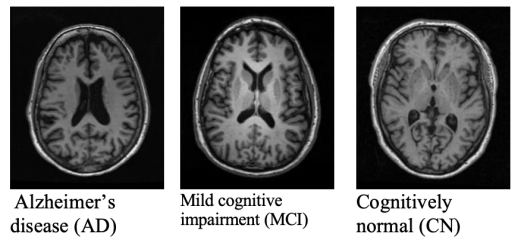

Alzheimer's
disease (AD)

Mild cognitive
impairment (MCI)

Cognitively
normal (CN)

Fig. 2: T1-weighted sMRI images of Alzheimer's disease (AD), mild cognitive impairment (MCI) and cognitively normal (CN) subjects. One can note the changes in the brain structure

We experimented with AlexNet to extract the features from 2D images. First, the top layers of the AlexNet network are adopted for the three classes. Then, the network is fine-tuned to extract the features. The selected features are scaled using a standard scaler and passed to the principal component analysis (PCA) for dimensionality reduction. The extracted features are finally passed to the next step for class decomposition.

For class decomposition, We use the elbow method to decide the optimal number of clusters for k-means clustering. After K-means was conducted with $k = 2$, the three class labels CN, MCI and AD are divided into six sub-classes. Table 2 shows the classes' distributions before and after class decomposition.

| Class | Instances | Class | Instances |
|-------|-----------|-------|-----------|
| CN | 704 | $CN_1$ | 10 |
| | | $CN_2$ | 694 |
| MCI | 720 | $MCI_1$ | 99 |
| | | $MCI_2$ | 621 |
| AD | 720 | $AD_1$ | 26 |
| | | $AD_2$ | 694 |

Table 2: Classes distribution before and after class decomposition

## 4.2 Classification and Class Composition

We aimed first to explore hyper-parameters for the model's training and determine which layers we should fine-tune. The performance of the ResNet50 and VGG19 networks was tested using the Adam algorithm to find the optimal number of learning rates within (0.01 and 0.001) over 200 epochs and batch sizes of 64. As shown in Table 3, when fine-tuning the top two layers, the VGG19 network achieved the highest performance with an accuracy of 98% while ResNet50 achieved 94%. However, the ResNet50 performs well when fine-tuning the top thirteen layers with an accuracy of 97%, while the VGG19 network showed a drop in performance when fine-tuning more layers. As shown in Table 3, the VGG19 model achieved the highest accuracy of 98% when training the two top layers on ADNI data, while model accuracy decreased to 94% and 96% as more layers were trained. We notice that transferring knowledge from some layers outperforms strict training on the target task. For instance, when fine-tuning the top two layers, the VGG19 performance improved and achieved the most outstanding outcomes. With Resnet50, on the other hand, we had to fine-tune more layers in order to get the best results, which demonstrates that knowledge transferability varies between networks and even between layers. Figure 3 illustrates the learning curve accuracy and loss for model training and testing obtained using VGG19.

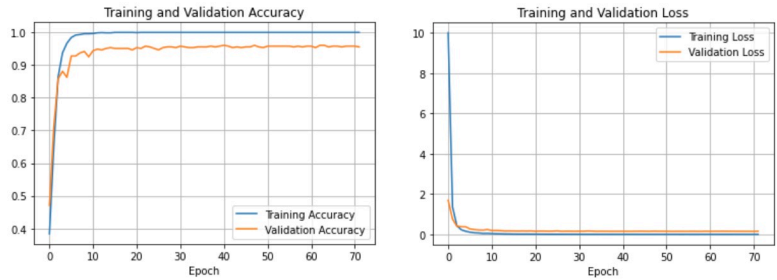

Fig. 3: The learning curve accuracy and loss error obtained by VGG19 pre-trained network

| Network | Learning Rate | Fine-tuned layers | Accuracy (%) | Specificity (%) | Sensitivity (%) |
|---|---|---|---|---|---|
| VGG19 | 0.01 | block5 _conv4 | 92 | 95 | 91 |
| | | dense | 96 | 97 | 95 |
| | 0.001 | block5_conv1 | 94 | 96 | 93 |
| | | block5_conv4 | 96 | 98 | 96 |
| | | dense | 98 | 99 | 98 |
| ResNet50 | 0.01 | conv5_block3_1_conv | 86 | 93 | 86 |
| | 0.001 | conv5_block3_1_conv | 97 | 98 | 96 |
| | | dense | 94 | 96 | 94 |

Table 3: Hyper-parameters selection for classification phase

## 4.3 Comparison with existing methods

Comparison with previous research findings has been challenging because studies differ in datasets, data preparation strategies and dimensional reduction methods and measurements. This section discusses the results in relation to other recent methods in terms of training size and accuracy. It is noteworthy that the methods' results are comparable, even though the studies may employ different experimental setups.

The training size was calculated based on the sample used in training these models. For example, our work used 2,680 images, which were divided into 80% for training and validation and 20% for classifier testing, resulting in a training size of 1,715. Table 4 reports the results of the binary and ternary classifiers of the model. To further validate our proposed architecture, we adapted the same architecture for a three-way binary classification by changing the final classification layer.

| No. | Training size | AD vs CN (%) | AD vs MCI (%) | MCI vs CN (%) | AD vs MCI vs CN (%) |
|---|---|---|---|---|---|
| [10] | 2,560 | 99.4 | 99.2 | 99.0 | 95.2 |
| [14] | 1,731 | 95.4 | 82.2 | 90.1 | 85.5 |
| [9] | – | 90.4 | 77.2 | 72.4 | |
| [12] | 37,590 | 95.9 | 99.3 | 96.8 | |
| | | 98.9 | 99.1 | 97.1 | |
| proposed model | 1,715 | 99.0 | 99.0 | 98.0 | 98.0 |

Table 4: Comparison of classification performance with state-of-art studies

We compared our model with four other state-of-the-art models; the results for these models are what is reported in their papers. The results in Table 4 shows that our model ternary classification outperforms the other approaches with an accuracy of 98%, a 3% improvement over the state-of-the-art performance, which archived 95.19%, 85.53%, and 89.47%. Furthermore, except for [10], our model's binary classification results outperformed the other approaches. Compared to this study, our model used a smaller training size (1,715 images) extracted from fewer subjects.

Our proposed model utilises transfer learning and class decomposition. Transfer learning deals with the challenge of the limited availability of annotated data while using class decomposition enhances model performance because it makes learning the class boundaries of a dataset uncomplicated and, as a result, can deal with any irregularities in data distribution. Using transfer learning with class decomposition leads to better accuracy than other state-of-the-art methods. Class decomposition makes learning the class boundaries of a dataset uncomplicated and deals with any irregularities in data distribution. It divided the Alzheimer's classes into six new subclasses, and this assisted in revealing the hidden patterns in each class and made more accurate predictions.

## 5 Conclusion

In this paper, we propose a model that integrates transfer learning with a class decomposition approach for diagnosing Alzheimer's from structural sMRI images. In addition, we use the entropy-based approach to select the training dataset that contains the most informative data. We use the VGG19 ImageNet-trained weights network to obtain highly accurate results. We compared our findings to those of four other cutting-edge procedures using the ADNI dataset. With an accuracy of 98.3%, our model ternary classification outperforms the other approaches, representing a 3% improvement over the state-of-the-art performance. In addition, except for [10], our model outperformed the others in the binary classification task.

For future work, we will conduct more analysis to investigate the impact of each component of our method (namely, class decomposition and the extracted features). Also, the two-dimensional slices have limitations in covering all of the brain's regions, therefore causing information loss. Other approaches for image segmentation can be considered in future work. Additionally, we aim to combine the model with other data modalities for the diagnosis of Alzheimer's disease, such as genomic data and Electronic Health Records. We seek to use the model to discover the hidden patterns in the mild cognitive impairment (MCI) category to reveal the conversion of mild cognitive impairment (MCI) patients to Alzheimer's disease by discriminating MCI levels based on cognitive decline. We also aim to extend the model architecture to be more scalable and also to include other factors that can assist in diagnosing Alzheimer's disease, such as the Mini-Mental State Examination (MMSE) score and age, which could increase model performance.